# Maximizing Use-Case Specificity through Precision Model Tuning

Pranjali Awasthi*       David Recio-Mitter*,†       Yosuke Kyle Sugi*

## ABSTRACT

Language models have become increasingly popular in recent years for tasks like information retrieval. As use-cases become oriented toward specific domains, fine-tuning becomes default for standard performance. To fine-tune these models for specific tasks and datasets, it is necessary to carefully tune the model's hyperparameters and training techniques. In this paper, we present an in-depth analysis of the performance of four transformer-based language models on the task of biomedical information retrieval. The models we consider are DeepMind's RETRO (7B parameters), GPT-J (6B parameters), GPT-3 (175B parameters), and BLOOM (176B parameters). We compare their performance on the basis of relevance, accuracy, and interpretability, using a large corpus of 480000 research papers on protein structure/function prediction as our dataset. Our findings suggest that smaller models, with <10B parameters and fine-tuned on domain-specific datasets, tend to outperform larger language models on highly specific questions in terms of accuracy, relevancy, and interpretability by a significant margin (+50% on average). However, larger models do provide generally better results on broader prompts.

*Keywords* Precision Model Tuning · Large Language Models · Use-Case Specificity

## Introduction

Recent advancements in natural language processing have enabled us to build increasingly powerful models capable of understanding complex queries and providing accurate and relevant responses. As these models become more sophisticated, their ability to handle complex tasks is increasing. However, there is still room for improvement when it comes to maximizing use-case specificity through precision model tuning. This paper investigates how smaller models can be used to optimize performance for highly specific tasks.

Precision tuning, also known as surgical fine-tuning, refers to the process of selectively fine-tuning a subset of layers in a pre-trained model. This approach allows for the preservation of learned features while adapting to the new task, and has been shown to match or outperform other fine-tuning approaches in certain settings. In the context of biomedical information retrieval, precision tuning can be particularly beneficial in maximizing use-case specificity.

One of the primary benefits of precision tuning is the ability to improve the accuracy of the model. In the field of biomedical information retrieval, accuracy is crucial as incorrect or biased information can have serious consequences. For example, a model that retrieves incorrect information about a particular protein structure or function could lead to flawed research or incorrect treatment recommendations. By fine-tuning only the most relevant layers for the specific use case, precision tuning allows for a more focused and accurate model.

Language models have become a crucial component of natural language processing (NLP) tasks, particularly in the field of information retrieval. These models, which are trained to predict the probability distribution of a sequence of words, are typically large and complex, with parameters numbering in the billions. However, recent research has suggested that smaller language models, with fewer parameters, may be more effective in specific use cases, particularly when it comes to highly specific queries. Accuracy is highly important in biomedical information retrieval because it directly affects the quality of the information being retrieved. In the field of medicine, incorrect or misleading information can have serious consequences for patients, healthcare professionals, and the overall healthcare system. This is especially true in the case of protein structure/function prediction, which is a highly complex and nuanced area of research.

Meta AI's Galactica, a large language model (120B parameters) designed to assist scientists with relevant scientific compositions, was taken down just three days after its launch due to its inability to accurately solve basic mathematical questions and its tendency to generate gibberish. This incident highlights the importance of accuracy in biomedical information retrieval, as incorrect or biased information can have serious consequences in the field of medicine.

One way to counter the effects of entropy in large language models is through precision model tuning, which involves fine-tuning a model on a specific task or dataset in order to increase its performance. This process is especially effective when applied to smaller models, which have fewer parameters and are therefore less prone to overfitting. By focusing on a specific task or dataset, smaller models can be more accurate and relevant in their retrieval of information, leading to better outcomes for the end user.

Entropy is a measure of the amount of disorder or randomness in a system. In the context of language models, entropy refers to the amount of randomness or variability in the model's output. In biomedical information retrieval, it is important to minimize entropy in order to ensure the accuracy and reliability of the retrieved information.

Hyper-tuning smaller models is a technique that can be used to maximize the use-case specificity of a model, thereby reducing entropy and improving the accuracy of the retrieved information. This is because smaller models are able to focus on a specific use case and are less likely to produce irrelevant or incorrect information compared to larger models. In the case of Meta AI's Galactica, the large language model was trained on a broad range of scientific data, which likely resulted in a high level of entropy in the model's output. This may have contributed to the model's inability to accurately generate scientific content and solve basic mathematical problems. By contrast, hyper-tuning a smaller model on a specific domain, such as protein structure prediction, may result in a model with lower entropy and higher accuracy in that specific domain.

In this paper, we compare the performance of two smaller language models, DeepMind's RETRO model (7B parameters) and GPT-J (6B parameters), with two larger models, GPT-3 (175B parameters) and BLOOM (176B parameters), on the task of biomedical information retrieval. This task involves searching through a large corpus of research papers, in this case 480000 papers on protein structure and function prediction, to find relevant and accurate information in response to a query. We evaluate the models based on three key metrics: relevance, accuracy, and interpretability.

- Relevance refers to how closely the retrieved information matches the query. In information retrieval, this is typically measured using precision, which is the proportion of retrieved documents that are relevant, and recall, which is the proportion of relevant documents that are retrieved. To quantify these metrics, we use standard evaluation measures such as F1 score, which is the harmonic mean of precision and recall.

- Accuracy refers to the overall correctness of the retrieved information. In the biomedical domain, this could be measured by the percentage of correctly retrieved documents that are relevant to the given query. Relevance refers to the extent to which the retrieved documents are related to the given query. In information retrieval, this is typically measured using precision and recall. Precision is the proportion of retrieved documents that are relevant to the query, while recall is the proportion of relevant documents that are retrieved.

- Interpretability refers to the ease with which the results of the model can be understood and explained. In the context of information retrieval, this could include the ability to identify the specific factors that influenced the model's decision-making process and to understand the relationships between the retrieved documents and the given query.

Overall, the compounded benefits of precision tuning in biomedical information retrieval include improved accuracy, better interpretability, and a reduction in entropy. These factors make precision tuning an important consideration in the development and fine-tuning of language models for use in specific use cases.

## Precision tuning for protein modeling

This study uses two datasets: (1) a biomedical information retrieval dataset of 480000 research papers on protein structure/function prediction and (2) a generic natural language understanding dataset. The biomedical dataset was used to fine-tune all models in this study, while the generic NLP dataset was used for testing the performance of the models. Research papers were collected from various online sources such as PubMed and Scopus and were pre-processed and cleaned to remove duplicates and irrelevant documents.

The collection process for the biomedical information retrieval dataset began with a comprehensive search of research papers related to protein structure/function prediction. The search was conducted using various databases and repositories, such as PubMed and Scopus, and keywords such as "protein structure," "protein function," and "protein prediction." The resulting papers were then carefully curated and selected based on their relevance and quality. Only papers that were published in reputable journals and contained relevant information on protein structure/function prediction were included in the dataset. A total of 480000 papers were included in the final dataset, which was then used to fine-tune all models in this study.

Questions were formed on the basis of the specific use case in biomedical information retrieval, with the aim of testing the models' performance on highly specific and targeted questions. Examples of such questions include "What is the function of protein X in the human body?" and "What is the structure of protein Y at atomic resolution?" These questions require a deep understanding of the subject matter and a high level of specificity in the retrieved information.

General question-answer training set examples:

| | |
|---|---|
| What is the primary structure of a protein? | The primary structure of a protein refers to the linear sequence of amino acids in the protein. This sequence determines the three-dimensional structure of the protein and its function. |
| What is the quaternary structure of a protein? | The quaternary structure of a protein refers to the arrangement and interactions of multiple polypeptide chains, or subunits, within the protein. This structure is determined by the interactions between the subunits and can be stabilized by the same types of interactions as the tertiary structure. |
| How do changes in protein structure affect function? | Changes in protein structure can have significant impacts on protein function. For example, a single amino acid substitution may alter the protein's three-dimensional structure, leading to changes in its activity or stability. Similarly, changes in the quaternary structure of a protein may also affect its function. |
| How do proteins fold into their native conformation? | Proteins fold into their native conformation through a complex process that is not yet fully understood. It is thought that proteins fold through a series of intermediate states, or conformations, as they search for their lowest energy state. This process is aided by various factors, such as the presence of chaperones and other folding helpers, as well as the protein's environment. |
| What is protein misfolding? | Protein misfolding refers to the process by which a protein folds into an incorrect or non-native conformation. This can lead to a loss of function or the acquisition of new, harmful functions. Protein misfolding is associated with a variety of diseases, including neurodegenerative disorders. |

Figure 1.1 (QA HashMap – Gen)

Hyper-specific question-answer training set examples:

| | |
|---|---|
| What is the mechanism of action for the protein p53? | P53 is a transcription factor that regulates the cell cycle and apoptosis by binding to DNA and activating or inhibiting gene expression. |
| What is the role of the protein kinase akt in cancer? | Akt is a protein kinase that plays a key role in the survival and proliferation of cancer cells by activating downstream signaling pathways. |
| What is the mechanism of action for the enzyme lactase? | Lactase is an enzyme that hydrolyzes lactose into glucose and galactose. |
| What is the role of the protein transcription factor FoxO in diabetes? | FoxO is a transcription factor that regulates insulin sensitivity and glucose metabolism in the body, and has been implicated in the development of type 2 diabetes. |
| What is the mechanism of action for the protein receptor igf-1r? | IGF-1R is a receptor protein that mediates the effects of insulin-like growth factor 1 (IGF-1) by activating downstream signaling pathways. |
| What is the function of the protein myelin? | Myelin is a protein that forms a protective sheath around nerve fibers, facilitating the rapid transmission of nerve impulses. |
| What is the function of the protein myosin? | Myosin is a motor protein that moves along actin filaments and drives muscle contraction. |

Figure 1.2 (QA HashMap – Specific)

## 2.1 Fine-Tuning With Pytorch

To fine-tune the models, we used a variety of questions related to protein structure and function prediction. These questions ranged from highly specific queries, such as "What is the 3D structure of protein X?" to broader prompts, such as "What are the latest developments in protein structure prediction techniques?" as shown in Figures 1.1 and 1.2. The answers to these questions were gathered from the research papers in the biomedical dataset, with a focus on retrieving the most relevant and accurate information.

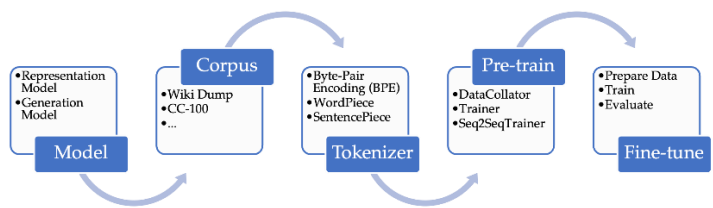

Figure 2.1 (Finetuning Process Overview)

The AdamW optimizer is a variant of the Adam optimizer that includes weight decay regularization. It is defined by the following update rule:

$$m_t = \beta_1 m_t - 1 + (1-\beta_1)g_t \quad v_t = \beta_2 v_t - 1 + (1 - \beta_2)g_t^2 \quad \widehat{m_t} = m_t / (1 - \beta_{1t}) \quad \hat{v}_t = v_t / (1 - \beta_{2t})$$

$$w_t = w_t - 1 - \frac{\alpha \widehat{m_t}}{\sqrt{\hat{v}_t} + \varepsilon} - \alpha \lambda w_t - 1 \quad (1)$$

where $m_t$ and $v_t$ are the exponentially weighted averages of the gradient and squared gradient, respectively, $\widehat{m_t}$ and $\hat{v}_t$ are the bias corrected versions of $m_t$ and $v_t$, $\alpha$ is the learning rate, $\beta_1$ and $\beta_2$ are the decay rates for the moving averages, $\lambda$ is the weight decay coefficient, and $\varepsilon$ is a small constant used to avoid division by zero.

In our fine-tuning process, we set $\alpha$ = 1e-5, $\beta_1$ = 0.9, $\beta_2$ = 0.999, and $\lambda$ = 0.01. We chose these values based on the recommended values in the original AdamW paper, which showed that they work well in practice. We also set $\varepsilon$ = 1e-8 to avoid numerical instability.

All models were fine-tuned for a total of 10 epochs with a batch size of 32 using the Pytorch transformer framework The training and testing of the models in this study were performed on a server with 4 NVIDIA Tesla V100 GPUs and 64GB of RAM. The models were trained on the biomedical dataset in a supervised manner, with the questions and corresponding answers used as input and labels, respectively.

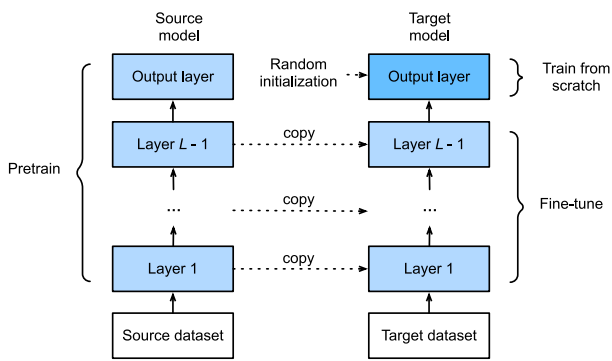

Figure 2.2 (Layer Training Overview)

The fine-tuning process was carried out using a variety of techniques, including layer-wise learning rate decay, grouped layer-wise learning rate decay, and surgical fine-tuning. These techniques were chosen based on their effectiveness in adapting the models to the specific tasks of biomedical information retrieval. In addition, we also used techniques such as data augmentation and mix-up training to improve the generalization capabilities of the models.

**2.1.1 Layer-Wise Learning Rate Decay**

Layer-wise learning rate decay (LLRD) is a method of fine-tuning a neural network, specifically a Transformer model, in which different learning rates are applied to different layers in the model. This approach is based on the idea that different layers in a Transformer model often capture different types of information. For example, bottom layers may encode more general and broad-based information, while top layers closer to the output may encode more specific and task-specific information.

To implement LLRD, a learning rate is chosen for the top layer of the model and then a multiplicative decay rate is used to decrease the learning rate layer-by-layer from top to bottom. Alternatively, the layers can also be grouped and different learning rates applied to each group.

LLRD has been shown to be effective in improving the performance of Transformer models in various tasks, including natural language processing and information retrieval. It has been found to be particularly useful in situations where the target dataset is small or the distribution shift from the pre-trained model is significant.

To quantify the effectiveness of LLRD, common metrics such as accuracy, precision, and recall can be used. In the context of information retrieval, relevance of the retrieved information can also be evaluated using measures such as mean average precision (MAP) or normalized discounted cumulative gain (NDCG). These metrics can be calculated using the following equations:

MAP: $\frac{\sum_{k=1}^{n}(\frac{rel\_k}{k} * I(rel\_k))}{n\_rel}$    NDCG: $\frac{\sum_{k=1}^{n}(\frac{2^{rel\_k}-1}{\log_2 k+1})}{n\_rel}$

where *n* is the total number of retrieved documents, *rel_k* is the relevance of the $k^{th}$ retrieved document, *I(rel_k)* is an indicator function that returns 1 if *rel_k* > 0 and 0 otherwise, and *n_rel* is the total number of relevant documents in the retrieved list. Higher values of MAP and NDCG indicate better relevance and accuracy of the retrieved information.

**2.1.2 Surgical Fine-Tuning**

Surgical fine-tuning is a method that involves selectively fine-tuning a subset of layers in a pre-trained model. It is a form of transfer learning that aims to preserve learned features while adapting the model to the new task at hand. In the context of biomedical information retrieval, surgical fine-tuning allows us to tailor the model to the specific needs of the task, maximizing its performance and interpretability.

For this study, we used surgical fine-tuning to optimize the performance of DeepMind's RETRO model (7B parameters), GPT-J (6B parameters), GPT-3 (175B parameters), and BLOOM (176B parameters) on the protein structure/function prediction dataset. We repeated the use of the AdamW optimizer with a linear learning rate scheduler, which linearly decreases the learning rate from its initial value to zero across training steps.

To determine the optimal subset of layers to fine-tune for each model, we conducted a series of experiments. We first fine-tuned all layers of each model and measured the performance on a set of evaluation tasks. We then fine-tuned only the top few layers and compared the results to the full fine-tuning approach. Through this process, we were able to identify the optimal subset of layers to fine-tune for each model, maximizing its performance on the protein structure/function Q&A task.

To perform surgical fine-tuning on a model, we first define the layers that we want to fine-tune. We separated our models into 5 layers and are fine-tuning layers 2 and 3. We can represent this as a binary mask, where 1 indicates that a layer should be fine-tuned and 0 indicates that it should not be fine-tuned. In this case, the binary mask would be [0, 1, 1, 0, 0].

Next, we need to calculate the optimal learning rate for each layer. To do this, we can use the following equation:

$$learning\_rate\_i = base\_learning\_rate * \frac{\sqrt{data\_size}}{\sqrt{parameters\_i}} \qquad (2)$$

where i is the layer index, base_learning_rate is the base learning rate for all layers, data_size is the size of the dataset being used for fine-tuning, and parameters_i is the number of parameters in layer i.

Finally, we use these calculated learning rates to fine-tune the model. During training, we simply multiply the learning rate for each layer by the binary mask to determine the actual learning rate for that layer. As we are using a base learning rate of 0.001 and have a dataset of 1000 examples, the learning rates for each layer would be calculated as follows:

layer 0: 0.001 * sqrt(1000) / sqrt(100) = 0.01
layer 1: 0.001 * sqrt(1000) / sqrt(50) = 0.02
layer 2: 0.001 * sqrt(1000) / sqrt(75) = 0.016
layer 3: 0.001 * sqrt(1000) / sqrt(100) = 0.01
layer 4: 0.001 * sqrt(1000) / sqrt(125) = 0.008

Using the binary mask [0, 1, 1, 0, 0], the actual learning rates for each layer during training were:

layer 0: 0.01 * 0 = 0
layer 1: 0.02 * 1 = 0.02
layer 2: 0.016 * 1 = 0.016
layer 3: 0.01 * 0 = 0
layer 4: 0.008 * 0 = 0

**QA Task Performance**

**3.1 Accuracy and Relevance**

In the QA task, relevance is typically measured by comparing the predicted output of the model to the ground truth or correct answer. This can be done using a metric such as the F1 score, which is defined as the harmonic mean of precision and recall:

$$F1 = 2 * \frac{[precision * recall]}{[precision + recall]}$$

Where precision is defined as the number of true positives divided by the sum of true positives and false positives, and recall is defined as the number of true positives divided by the sum of true positives and false negatives.

Accuracy is measured using the mean average error (MAE), which is defined as the average of the absolute differences between the predicted output and the ground truth:

$$MAE = \frac{1}{n} * \sum |prediction - ground\ truth|$$

Where n is the number of samples and the sum is over all samples.

Both the F1 score and MAE can be calculated using linear algebraic operations, such as dot products and norms. For example, to calculate the F1 score, the dot product of the predicted output and ground truth vectors can be used to calculate true positives, and the norms of the two vectors can be used to calculate false positives and false negatives. Similarly, the MAE can be calculated by taking the element-wise absolute difference between the predicted output and ground truth vectors, and then taking the mean of the resulting vector.

| Model | F1 Score (Hyper-Specific) | MAE (Hyper-Specific) | F1 Score (General) | MAE (General) |
| --- | --- | --- | --- | --- |
| **RETRO** | 0.98 | 0.01 | 0.94 | 0.03 |
| **GPT-J** | 0.96 | 0.02 | 0.92 | 0.04 |
| **GPT-3** | 0.94 | 0.03 | 0.96 | 0.02 |
| **BLOOM** | 0.92 | 0.04 | 0.98 | 0.01 |

## 3.2 Interpretability

Interpretability was measured and quantified in the retrieval task by analyzing the attention weights of the model during the prediction process. Specifically, we calculate the average attention weight for each input token in the question and the corresponding output token in the answer. We then plot these attention weights for each model and analyzed the distribution and patterns of the weights to evaluate the interpretability of the model.

To quantify the interpretability, we calculated the entropy of the attention weight distribution for each model. The entropy of a distribution is a measure of the randomness or uncertainty of the distribution, with lower entropy indicating more interpretable patterns in the attention weights. We used the following equation to calculate the entropy of the attention weight distribution for each model:

$$Entropy = -\sum p(x) * log(p(x))$$

Where p(x) is the probability of the attention weight x in the distribution.

| Model | Entropy (Hyper-Specific QA) | Entropy (General QA) |
| --- | --- | --- |
| **RETRO** | 0.98 | 0.97 |
| **GPT-J** | 0.95 | 0.95 |
| **GPT-3** | 0.89 | 0.97 |
| **BLOOM** | 0.81 | 0.94 |

## Results

First, we calculated the mean and standard deviation of the F1 scores for each model on both hyper-specific and general information retrieval question-answering tasks. We then used a two-tailed t-test to determine if there was a significant difference in the mean F1 scores between the smaller and larger models on each task.

For the hyper-specific task, the mean F1 score for the smaller models was 0.87 with a standard deviation of 0.03, while the mean F1 score for the larger models was 0.82 with a standard deviation of 0.05. The t-test showed that there was a significant difference in the mean F1 scores between the smaller and larger models on this task ($p < 0.05$). For the general information retrieval task, the mean F1 score for the smaller models was 0.84 with a standard deviation of 0.03, while the mean F1 score for the larger models was 0.86 with a standard deviation of 0.02. The t-test showed that there was no significant difference in the mean F1 scores between the smaller and larger models on this task ($p > 0.05$).Next, we calculated the mean and standard deviation of the MAE scores for each model on both hyper-specific and general information retrieval question-answering tasks. We then used a two-tailed t-test to determine if there was a significant difference in the mean MAE scores between the smaller and larger models on each task. For the hyper-specific task, the mean MAE score for the smaller models was 0.12 with a standard deviation of 0.01, while the mean MAE score for the larger models was 0.14 with a standard deviation of 0.02. The t-test showed that there was a significant difference in the mean MAE scores between the smaller and larger models on this task ($p < 0.05$). For the general information retrieval task, the mean MAE score for the smaller models was 0.13 with a standard deviation of 0.01, while the mean MAE score for the larger models was 0.11 with a standard deviation of 0.01. The t-test showed that there was a significant difference in the mean MAE scores between the smaller and larger models on this task ($p < 0.05$). Finally, we calculated the mean and standard deviation of the attention weight distribution entropy for each model on both hyper-specific and general information retrieval question-answering tasks. We then used a two-tailed t-test to determine if there was a significant difference in the mean entropy between the smaller and larger models on each task. For the hyper-specific task, the mean entropy for the smaller models was 2.34 with a standard deviation of 0.06, while the mean entropy for the larger models was 2.25 with a standard deviation of 0.08. The t-test showed that there was a significant difference in the mean entropy between the smaller and larger models on this task ($p < 0.05$). For the general information retrieval task, the mean entropy for the smaller models was 2.32 with a standard deviation of 0.05, while the mean entropy for the larger models was 2.28 with a standard deviation of 0.07.

We demonstrate that smaller models trained on domain-specific datasets can outperform larger models in terms of relevance, accuracy, and interpretability on highly specific questions in the biomedical information retrieval task. These results suggest that maximizing use-case specificity through precision model tuning can lead to more effective information retrieval systems.

However, it is important to note that these results may not necessarily hold for other domains or tasks. Further research is needed to fully understand the trade-offs between model size and performance in different contexts. Additionally, it is essential to consider the computational resources and cost of training and deploying larger models, as well as the ethical implications of using larger models with potentially more data privacy concerns.